\documentclass[letterpaper, 10 pt, conference]{ieeeconf}  

\IEEEoverridecommandlockouts                              %
\usepackage[table, dvipsnames]{xcolor}

\usepackage{graphicx}
\usepackage{amsmath}
\usepackage{amssymb}
\usepackage{booktabs}
\usepackage{mathtools}
\usepackage[accsupp]{axessibility}  
\usepackage{color}
\usepackage{array}
\usepackage{pdflscape}
\usepackage[longtable]{multirow}
\usepackage{longtable}
\usepackage{tabularray}
\usepackage{booktabs}
\usepackage{times}
\usepackage{epsfig}
\usepackage{amsmath}
\usepackage{bbm}
\DeclareMathAlphabet\mathbfcal{OMS}{cmsy}{b}{n}

\usepackage{float}
\usepackage{comment}
\usepackage{lipsum}
\usepackage{stfloats}
\usepackage{multicol}
\usepackage{multirow}
\usepackage{bm}
\usepackage{etoolbox}
\usepackage{icomma}
\usepackage{array}
\usepackage{tabulary}
\usepackage{xcolor}
\usepackage{paralist}
\usepackage{booktabs}
\usepackage{adjustbox}
\usepackage[small]{caption}
\usepackage{subcaption}
\usepackage{arydshln}

\definecolor{gray}{rgb}{0.3,0.3,0.3}
\definecolor{blue}{rgb}{0,0.5,1}
\definecolor{mask_red}{rgb}{1,0,0.8}
\definecolor{green}{rgb}{0.2,1,0.2}
\definecolor{rblue}{rgb}{0,0,1}
\definecolor{lightblue}{HTML}{6495ed}
\definecolor{lightred}{HTML}{F19C99}

\definecolor{graytablerow}{gray}{0.6}

\usepackage{pifont}
\usepackage{tabu}

\usepackage[utf8]{inputenc}
\usepackage[T1]{fontenc}
\usepackage{soul}
\usepackage{url}
\definecolor{bluegray}{rgb}{0.4, 0.6, 0.8}
\usepackage[pagebackref,breaklinks,colorlinks,citecolor=bluegray]{hyperref}
\usepackage[utf8]{inputenc}
\usepackage{graphicx}
\usepackage{amsmath}
\usepackage{booktabs}
\usepackage{algorithm}
\usepackage{algorithmic}
\usepackage[switch]{lineno}
\usepackage{cleveref}
\usepackage{pgfplots}
\usepackage{pgfplotstable}
\pgfplotsset{compat=1.5}
\usepackage{xcolor}
\usetikzlibrary {shapes.geometric}
\usepackage{color}
\usepackage{colortbl}

\usepackage{amsthm}

\theoremstyle{definition}
\newtheorem{definition}{Definition}[section]

\makeatletter
\patchcmd{\@makecaption}
  {\scshape}
  {}
  {}
  {}
\makeatletter
\patchcmd{\@makecaption}
  {\\}
  {.\ }
  {}
  {}
\makeatother

\usepackage{stfloats}

\title{\Large \bf
Comb, Prune, Distill: Towards Unified Pruning for Vision Model Compression
}
\author{Jonas Schmitt$^{*}$, Ruiping Liu$^{*}$, Junwei Zheng, Jiaming Zhang$^\dag$ and Rainer Stiefelhagen
\thanks{This work was supported in part by Helmholtz Association of German Research Centers, in part by the MWK through the Cooperative Graduate School Accessibility through AI-based Assistive Technology (KATE) under Grant BW6-03. This work was partially performed on HoreKa and Haicore.}
\thanks{Authors are with Institute for Anthropomatics and Robotics, Karlsruhe Institute of Technology, Karlsruhe, Germany.}
\thanks{Code is publicly available at: \href{https://github.com/Cranken/CPD}{https://github.com/Cranken/CPD}}
\thanks{$^{*}$Equal contribution.}
\thanks{$^{\dag}$Correspondence: jiaming.zhang@kit.edu}
}
\begin{document}

\maketitle

\begin{abstract}
Lightweight and effective models are essential for devices with limited resources, such as intelligent vehicles. Structured pruning offers a promising approach to model compression and efficiency enhancement. However, existing methods often tie pruning techniques to specific model architectures or vision tasks. To address this limitation, we propose a novel unified pruning framework \textbf{Comb, Prune, Distill (CPD)}, which addresses both model-agnostic and task-agnostic concerns simultaneously. Our framework employs a combing step to resolve hierarchical layer-wise dependency issues, enabling architecture independence. Additionally, the pruning pipeline adaptively remove parameters based on the importance scoring metrics regardless of vision tasks. To support the model in retaining its learned information, we introduce knowledge distillation during the pruning step. Extensive experiments demonstrate the generalizability of our framework, encompassing both convolutional neural network (CNN) and transformer models, as well as image classification and segmentation tasks. In image classification we achieve a speedup of up to $\times4.3$ with a accuracy loss of $1.8\%$ and in semantic segmentation up to $\times1.89$ with a $5.1\%$ loss in mIoU.

\end{abstract}

\section{Introduction}
\label{sec:intro}
\begin{figure}[!t]
\centering
    \input{figures/figure_results}
    \caption{\textbf{Model compression results.} (a) For image classification, our CPD method achieves ${\times}2.15$ speedup over ResNet-50. (b) For semantic segmentation on ADE20K, our method reduces ${\sim}48\%$ and ${\sim}26\%$ latency of ViT-DeiT-S and SeaFormer-L, respectively.    
    }
    \label{fig:pruning_results}
    \vskip -2ex
\end{figure}

The surge in artificial intelligence popularity largely stems from large-scale models in data centres~\cite{rameshZeroShotTexttoImageGeneration2021,achiam2023gpt}, but many scenarios lack such extensive computing resources. 
These limitations, due to space or energy constraints in autonomous vehicles~\cite{ma2021densepass,xiang2019importance,zhang2023cmx} in the intelligent transportation systems (ITS), assistive technologies~\cite{zhang2021trans4trans,liu2023open} and robotics~\cite{zheng2023materobot} or the desire for efficiency in unconstrained environments to reduce latency, necessitate more resource-efficient models. 
To alleviate this issue, various model compression approaches have been proposed. 
One promising approach is network pruning~\cite{hassibiOptimalBrainSurgeon1993,NIPS1989_6c9882bb}, which works by removing parameters from a well-trained model while preserving its accuracy. 
Numerous works are dedicated to the pruning of specific models~\cite{molchanovImportanceEstimationNeural2019,yangGlobalVisionTransformer2023}, such as for convolution-based or transformer-based models.
However, the common challenge associated with these methods is the over-reliance on a specific underlying network architecture or family, which in turn limits their generalizability to other model types. 
Expanding these methods to other architectures is a time-consuming endeavor that requires a deep understanding of the underlying structures.

To tackle this problem, we propose a unified model compression framework called \textbf{CPD} that includes three steps: (1) \textbf{Combing}; (2) \textbf{Pruning}; (3) \textbf{Distillation}. Specifically, it employs a combing process prior to pruning, which incorporates a \textit{task- and model-agnostic} algorithm for resolving dependencies on architecture. 
The design of the combing step is crucial as it can automatically extract structural information about any given model, so as to ensure the whole framework as model-agnostic. Furthermore, our framework removes the need for experts to manually define the dependency structure for new models which, in some cases, can be extremely time consuming. The extracted information from the combing step contains the dependency structure between layers which is needed for the next step, i.e., pruning. To this end, we present a new algorithm which uses the internal structure of the model to find dependencies between the input and outputs of the layers. Additionally, some more complex structures such as Transformer blocks introduce additional constraints which our algorithm can also automatically take into account.

While the combination of combing and pruning compress the model by removing redundant parts, Knowledge Distillation (KD) serves as an additional technique for enhancing the performance of a pre-existing compact architecture. The models with superior performance are typically resource-intensive, resulting in higher latency, whereas compact models are efficient but may exhibit lower effectiveness. KD methods are used to transfer the knowledge from an effective but cumbersome model (teacher) to a compact one (student). In our case, we use the original model as the teacher which guides the pruned model as the student during our pruning process. To maximize the performance gains from KD for pruned models, we systematically investigate various KD techniques, compare their effects on the pruning process, and find the best solution for our framework.

To verify the model-agnostic nature of our CPD framework, we apply it to two widely used architectures, CNNs~\cite{he2016resnet} and Transformers~\cite{wan2023seaformer,dosovitskiy2021vit}. Additionally, we conduct extensive experiments on two diverse datasets: ImageNet~\cite{dengImageNetLargescaleHierarchical2009} for image classification, and ADE20K~\cite{zhou2017ade20k} for semantic segmentation, to evaluate the task-agnostic capabilities of CPD. Our proposed method achieves a speedup of up to ${\times}4.31$ on ResNeXt-50 and ${\times}2.15$ on ResNet-50 (Fig~\ref{fig:pruning_results}) in image classification. 
In semantic segmentation, our method obtains reductions in latency of ${\sim}48\%$ and ${\sim}26\%$ with losses in mIoU of $5.1\%$ and $1.7\%$ for ViT-DeiT-S and SeaFormer-L respectively, as shown in Fig.~\ref{fig:pruning_results}.

Our main contributions can be summarized as follows:
\begin{compactitem}
    \item[(1)] We propose a unified model compression framework: Combing, Pruning and Distillation. Our combing step can extract the dependency structure of any given architecture, enabling the pruning to be fully model-agnostic. 
    \item[(2)] We investigate the combination of pruning and knowledge distillation to further improve the results of the model after pruning and finetuning. 
    \item[(3)] We compare our pruned models to their baselines and show that our method can improve the efficiency of the models with an acceptable trade off in predictive performance. For example, our method achieves a speedup of over $\times2$ on ResNet-50. 
\end{compactitem}

\section{Related Work}
\label{related_work}

\subsection{Model-agnostic Pruning}
Manually defining the pruning structure is time-consuming for complex models and lacks transferability to new architectures.
To solve this issue, Li et al.~\cite{liPruningFiltersEfficient2017} propose a method to prune CNN-based architectures including residual networks with skip connections. Similarly, Liu et al.~\cite{liuGroupFisherPruning2021} show another method for pruning CNNs by using mask sharing for coupled channels. Their methods support group-/depth-wise convolutions with DFS identifying coupled layers. 
For Transformer-based architectures, Yang et al.~\cite{yangGlobalVisionTransformer2023} present a pruning scheme for vision transformers. They focus on pruning the structures of a ViT model independently while keeping the dimensions between the structures consistent. 
Fang et al.~\cite{fangDepGraphAnyStructural2023} propose a model-agnostic framework for general structural pruning of arbitrary networks. 
Their solution still requires manual effort for novel architectures. In this paper we therefore propose a layer-wise dependency resolving algorithm which uses no prior knowledge to generate a complete pruning scheme. This allows us to prune novel architectures while showing good results in respective tasks.

\subsection{Knowledge Distillation Assisted Pruning}

The previous work~\cite{aghli2021combining} proposes a network-compression method
for convolutional architectures by combining weight pruning and knowledge distillation (KD). In \cite{bai2022dynamically}, a two-stage KD method is adopted to perform model compression on SegFormer~\cite{xie2021segformer} models for semantic segmentation. 
The HomoDistil method~\cite{liangHomoDistilHomotopicTaskAgnostic2023} further investigate the use of KD in combination with iterative neuron pruning for transformer architectures. In addition to the student's task loss, they propose a combination of distillation losses between student and teacher. 
In this work, we further explore the effectiveness of using KD in unified pruning methods, i.e., task-agnostic and model-agnostic at the same time during pruning.

\begin{figure}
  \centering
   \includegraphics[width=1\linewidth]{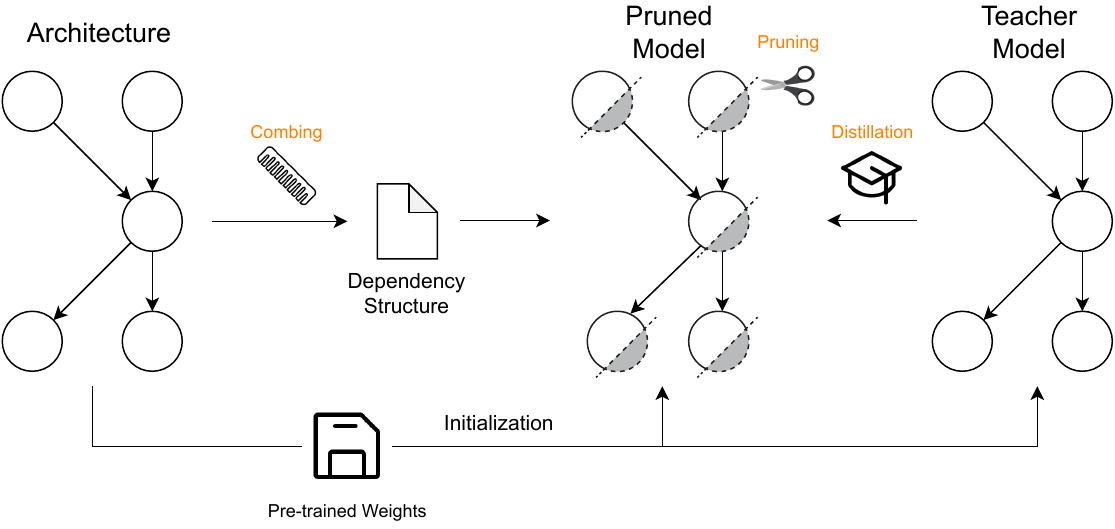}

   \caption{\textbf{Overview of CPD pipeline} including Combing, Pruning, Distillation. In the combing step (Sec. \ref{sec:meth_combing}), our dependency resolving algorithm extracts the dependency structure of the given architecture. Afterwards we initialize the to be pruned model (student) and the original model (teacher) with the same weights and start pruning (Sec. \ref{sec:meth_pruning}) the model. While pruning, we use KD (Sec. \ref{sec:meth_kd}) to help the student to retain more information.}
   \label{fig:overview}
   \vskip -2ex
\end{figure}
\section{Methodology}
\label{methodology}

\subsection{Framework Overview}

An overview of our method is given in \Cref{fig:overview}. Our method relies on the combination of three components. As mentioned before, it is needed to ensure matching dimensions of the inputs to specific operations in the model such as \textit{additions} and \textit{multiplications}. For this purpose, we introduce a layer-wise dependency resolving algorithm designed to detect such dependencies. The algorithm produces a set of coupling groups that include the parameters requiring simultaneous pruning to maintain coherent channel dimensions. 

Based on these coupling groups, we can start pruning a given model. Instead of randomly picking a group and pruning the contained neurons, we use a Hessian-based importance score to rank the neurons according to their importance. At each iteration, we remove the least important neurons from the model. To assist with the pruning and retention of predictive performance, we also investigate the use of KD in combination with the task-specific loss of the model. We detail the three major steps in the following. 

\subsection{
Combing Pipeline
}\label{sec:meth_combing}

\begin{figure}[t]
  \centering
   \includegraphics[width=0.8\linewidth]{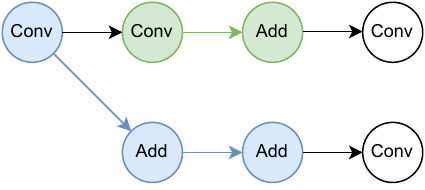}
   \caption{\textbf{Example of direct relation} between operations in a model. Operations with the same color are directly related}
   \label{fig:direct-relation}
   \vskip -2ex
\end{figure}

Firstly, we formalize a model $\phi$ consisting of its set of $n$ operations $f_i$ as $\phi=\{f_i\}_{i=0,\dots,n}$.
To find the output-input dependencies between layers of a model, we need to define the direct relation between two operations $f_i,f_j\in\phi$.

\begin{definition}[Direct Relation]
    $\delta(f_i,f_j)=1\leftrightarrow f_i$ and $f_j$ are directly connected without a stop operation in between.
\end{definition}
A stop operation in this context is an operation that may change the size of the channel dimensions between its input and output, e.g. linear layers or convolution layers. 
Note that operations such as reshaping are not regarded as stop operations, because they don't alter the size but only change the layout of the channel dimensions. 
Figure~\ref{fig:direct-relation} shows an example of direct relations in a model.

Another type of operation to consider in a network is what we define as coupling operations. The main characteristic of coupling operations is that they take multiple inputs and require the inputs to have a matching channel dimension in at least one of their channels. 
The coupling operations ``couple" the two inputs, whose channel dimensions must be kept consistent when pruning, so the previous layers of the inputs are supposed to be pruned together. 
Common examples of such operations are additions or matrix-matrix multiplications. We denote the set of all coupling ops in a model $\phi$ as $cpl(\phi)$.

Based on these definitions we can therefore find all the directly following coupling operations $c(w_i)$ for a given stop operation $w_i$ by:
\begin{equation}
    c(w_i)=\{f_j\mid f_j\in cpl(\phi), \delta(w_i,f_j)\}.
\end{equation}

\begin{figure}[t]
  \centering
   \includegraphics[width=0.8\linewidth]{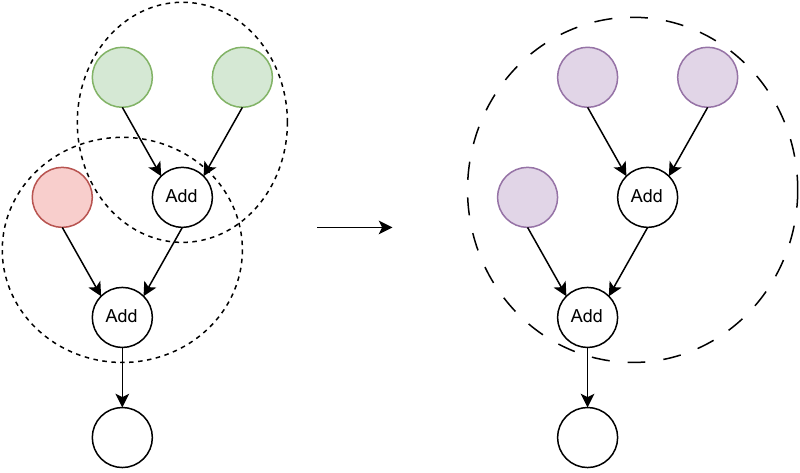}

   \caption{\textbf{Merging of coupled subgroups} based on their common parent coupling operation}
   \label{fig:layer-grouping}
   \vskip -2ex
\end{figure}

Finally, to find the final coupling groups $g_i$ in the model we accumulate all operations that share at least one common coupling operation as illustrated in Figure~\ref{fig:layer-grouping}, 
i.e.:  
\begin{equation}
    g_i=\cup\{\{v,w\}\in \phi \times \phi \mid c(v)\cap c(w)\neq\emptyset\}.
\end{equation}

At this point, these coupling groups solely serve to specify which operations' output channels should be pruned together at a given pruning step.
However, each of the operations in a coupling group also has a set of directly following operations. These following operations depend on the output channel dimensions of their previous operations, resulting in an inherent output dependency. Therefore, we must prune the input channel dimensions of the following operations as well to ensure consistency between inputs and weights of layers.

\subsection{Pruning Pipeline}\label{sec:meth_pruning}

Once our dependency-resolving algorithm identifies the coupling groups, the pruning process of the model can be started.
Firstly, we should decide what operations and which channels to remove at a given pruning step.
For this purpose, 
we employ a Hessian-based importance scoring method that quantifies the significance of a parameter or channel to the predictive performance of the model.
To preserve performance while reducing the model size, we enhance the pruning process by incorporating KD, using the unpruned model as the teacher.

\noindent\textbf{Importance Scoring.}
While removing parameters randomly from a model may lead to some success~\cite{liuUnreasonableEffectivenessRandom2022}, adopting a systematic parameter ranking and selection strategy proves to be a more effective approach.
Such strategies quantify the importance of parameters with a given model. To this end, multiple importance scoring methods have been proposed~\cite{heChannelPruningAccelerating2017,lin2017refinenet,linHRankFilterPruning2020,liuLearningEfficientConvolutional2017}. In this work, we choose to use a greedy strategy based on Hessian importance scoring. Note that our pruning method is independent of the actual importance score. We propose a Hessian-based importance function due to its effectiveness in pruning. Research has shown that a flatter curvature profile of the loss function is more resistant to small perturbations of the input which are introduced by pruning~\cite{moosavi-dezfooliRobustnessCurvatureRegularization2018,yangGlobalVisionTransformer2023}. Coincidentally, the curvature profile of the loss function corresponds to the set of eigenvalues of the Hessian matrix ~\cite{moosavi-dezfooliRobustnessCurvatureRegularization2018}. For a given loss function $\mathcal{L}$, the Hessian matrix is defined as
\begin{equation}
    H_{i,j}=\frac{\delta^2 \mathcal{L}}{\delta x_i,\delta x_j},
\end{equation}
where $w_i, w_j$ represent elements of the model weight $\textbf{w}$. 
Since we only need the eigenvalues of the Hessian matrix, or more specifically, the Frobenius norm of the eigenvalues $\sum_i \lambda_i^2$, we can skip the difficult computation of the Hessian matrix and instead calculate 
\begin{equation}
    \sum_i \lambda_i^2 = \mathbb{E} ||Hz||^2, z \sim N(0,I_d),
\end{equation}
where $z$ is a random vector in the normal distribution~\cite{hutchinsonStochasticEstimatorTrace1989}.

Rather than directly calculating $Hz$, we can further approximate the above equation with a finite difference $Hz\approx\frac{\triangledown(x+hz)-\triangledown \mathcal{L}(x)}{h}$, where $h$ is a small positive constant. Additionally, instead of directly pruning a given weight $w$ we instead introduce a binary mask $g_w$ to every prunable parameter in the model where the mask is initialized as $g_w=1$. This leads us to the gated weights $G_w=g_w w$.
Finally, we can define importance score for binary gates $g_w$ and the channels of the corresponding weight as:
\begin{align}
\begin{split}
    I(g_w )&=\mathbb{E}_z ||(\triangledown_{g_w} \mathcal{L}(g_w+hz)-\triangledown_{g_w} \mathcal{L}(g_w))/h||^2 \\
&= \mathbb{E}_z ||((1+hz) \triangledown_w \mathcal{L}(w)w-\triangledown_w \mathcal{L}(w)w)/h||^2 \\
&= \mathbb{E}_z ||(hz\triangledown_w \mathcal{L}(w)w)/h||^2 \\
&=(\triangledown_w \mathcal{L}(w)w)^2 \mathbb{E}_z ||z||^2 \\
&=(\triangledown_w \mathcal{L}(w)  w)^2.
\end{split}
\end{align}
Here we use the relation  $\triangledown_{g_w} \mathcal{L}(g_w)=\frac{\delta\mathcal{L}}{\delta G_w}\frac{\delta_{G_w} G_w}{\delta g_w }=\mathcal{L}(w)w$ to further simplify the equation.
Due to the availability of the weights and their gradients during the backpropagation pass, this importance score can be calculated in an efficient manner.
Additionally, we introduce a regularization parameter $R$, which aims to normalize the importance based on the expected performance impact of pruning. Since we can’t easily precompute the exact impact, we use a proxy score instead. 
Since memory reduction correlates with network speedup more directly compared to other metrics like FLOPs~\cite{liuGroupFisherPruning2021}, we use memory normalization specifically. 

To obtain the memory complexity for removing a single channel of a given weight or rather operation, we calculate $m_{single}=b\times h\times w$ where $b$ is the batch size, $h$ and $w$ are the height and width of the output tensor respectively. To calculate the total memory complexity for the whole masked parameter $G_w$, we calculate $m_{total,w}=m_{single}\times\sum_i g_w$ . Since $g_w$ is a binary mask, only the remaining unpruned parameters will affect the memory complexity. Therefore, the final importance score is given by 
\begin{equation}
    I_R(G_w)=\frac{I(G_w)}{m_{total,w}}.
\end{equation}

\noindent\textbf{Pruning the Model.}
With a method in place for ranking the importance of neurons within a model, we can mask them gradually until we have reached the desired overall threshold of pruned parameters. With this aim, we employ a greedy strategy to prune parameters at a fixed interval with $n$ training iterations. 
During this interval, we calculate the importance score for every neuron in the backpropagation pass and accumulate the scores over the whole interval. Since we are not only pruning individual neurons but groups of neurons instead, we aggregate the individual scores at the end of the iteration to calculate an importance score for the whole group. Finally, we remove the least important, i.e., the lowest importance score, group of neurons at the pruning interval by masking them.

\subsection{Knowledge Distillation}\label{sec:meth_kd}
In addition to using the normal task loss, we use knowledge distillation (KD) to assist the pruning and fine-tuning process.
The idea behind KD is to use a larger teacher to guide a smaller student model during training. In its simplest form, this is done by trying to match the output of the student to the output of the teacher. Specifically, when given a teacher model $\phi^T$, a student model $\phi^S$ and their respective output class probability distributions $f(\phi^\cdot)\in \mathbb{R}^C$, our objective is to optimize the following equation:
\begin{equation}
\label{eqn:std_kd}
    \min_{\phi^S} \mathcal{L}(\phi^S)+\mathcal{L}_{KL}(f(\phi^T),f(\phi^S)),
\end{equation}
where $\mathcal{L}(\phi^S)$ is the loss and $\mathcal{L}_{KL}(f(\phi^T),f(\phi^S))$ is the KL-Divergence between the student’s and the teacher’s output class prediction probability distributions~\cite{hintonDistillingKnowledgeNeural2015}.

Following the simple form of vanilla KD, various logit-based KD frameworks have been introduced to align the teacher and student model outputs while eliminating structural redundancy.
We select some of these works to investigate their applicability and generalizability when combined with our proposed pruning pipeline.
This means that we strictly focus on model-agnostic KD methods and exclude e.g., transformer-specific KD methods. 
In contrast, we include task-specific KDs in our experiments because it does not affect the generalizability of our pruning pipeline.

\noindent\textbf{Channel-wise Knowledge Distillation (CWD)} is for dense prediction tasks~\cite{shuChannelwiseKnowledgeDistillation2021}. The main difference of CWD compared to the basic KL-based distillation is to minimize the KL distance between the channel-wise output probability distributions. 
Specifically, instead of only calculating the KL distance between each class probability, CWD calculates the KL distance at each pixel in a channel-wise manner.

We denote the output activation probability distributions, i.e. feature activation maps, as $h(\phi_{c,i}^\cdot)\in\mathbb{R}^{C\times (H\times W)}$, where $ c=1,\dots, C$ is the channel index and $i=1,\dots, H\times W$ the spatial position. Thus, the definition of the CWD loss is:
\begin{equation}
\label{eq:cwd}
\mathcal{L}_{CWD}=\frac{\mathcal{T}^2}{C} \sum_{c=1}^C \sum_{i=1}^{H\times W} h(\phi^T_{c,i}) \cdot \log (\frac{h(\phi^T_{c,i})}{h(\phi^S_{c,i})}),
\end{equation}
where $\mathcal{T}$ is the same temperature hyperparameter as used to calculate the output probability distributions. $\mathcal{L}_{CWD}$ is then used to replace the standard $\mathcal{L}_{KL}$ in Eq.~\ref{eqn:std_kd}. 
The authors hypothesize that, due to the asymmetry of the KL divergence used in Eq.~\ref{eq:cwd}, 
the student is led to more effectively minimize the information loss in cases where $h(\phi^T_{c,i})$ is large, i.e. in high saliency regions of the feature activation map.
In our dense prediction experiments, we find that using CWD leads to better results than the standard KL formulation which also supports this hypothesis.

\noindent\textbf{Cross-Image Relational Knowledge Distillation} is another KD method specifically designed for semantic segmentation~\cite{yangCrossImageRelationalKnowledge2022}. The main idea of the authors is to introduce cross-image relational knowledge to the KD process. This method augments the standard KD form given by Eq.~\ref{eqn:std_kd} by three additional losses. First, they introduce a mini-batch pixel-to-pixel loss $\mathcal{L}_{batch\_p2p}$ which aims to implement a pixel-wise alignment among the output feature maps within the same mini-batch. This is accomplished by guiding the student to learn the pair-wise cross-image relational information generated by the teacher for all images within one mini-batch. 
In particular, our goal is to ensure that the student's pairwise similarity matrices of the feature maps closely match those generated by the teacher.

Since $\mathcal{L}_{batch\_p2p}$ captures relationships only within the confined mini-batch, the authors introduce a second type of loss: the memory-based pixel-to-pixel loss, denoted as $\mathcal{L}_{memory\_p2p}$.
This component relies on a global class-aware memory queue across all mini-batches to involve global dependencies among pixels. 
For each output embedding the student and teacher produce, they sample a specified number of embeddings from the memory queue. 
Then, similar to $\mathcal{L}_{batch\_p2p}$ the similarity matrices are calculated for the student and teacher separately. Finally, the student is guided to mimic the teacher's similarity matrices by adjusting the student to minimize the KL divergence between matrices.

We found that discrete pixel embeddings may be not enough to fully capture the complex relations inside the images, and introduce a third term, the memory-based pixel-to-region loss $\mathcal{L}_{memory\_p2r}$. The formulation is similar to that of $\mathcal{L}_{memory\_p2p}$, with the distinction that the memory queue now stores region embeddings instead of pixel embeddings.
Collectively, we derive the following equation for the loss:
\begin{equation}
    \begin{split}
    \mathcal{L}_{CIRKD} & =\mathcal{L} + \mathcal{L}_{KL} + \alpha \mathcal{L}_{batch\_p2p} \\
     & +\beta \mathcal{L}_{memory\_p2p} + \gamma \mathcal{L}_{memory\_p2r},
    \end{split}
\end{equation}
where $\alpha, \beta, \gamma \in [0.1, 1]$.

\section{Experiments}
\label{experiments}

\subsection{Settings}
\noindent \textbf{Implementation Details.} 
The models used in our experiments are off-the-shelf models from MMPretrain~\cite{mmpretrain2023} and MMSegmentation~\cite{mmseg2020}. All trainings and evaluations were done on NVIDIA GeForce RTX 2080 Ti GPUs. To train and finetune the models, we used the original configurations as given by the authors of the respective models.

\noindent \textbf{Datasets.} 
We conduct experiments on the ImageNet~\cite{dengImageNetLargescaleHierarchical2009} dataset to evaluate the performance on different architectures.
To prove our pruning method on dense prediction tasks like semantic segmentation, we use the ADE20K~\cite{zhou2017ade20k} dataset. It comprises 150 diverse classes of objects and stuff. It is split up in over 20K training, 2K validation and 3K test images. Furthermore, we also conduct our ablation studies on KD in the pruning process with this dataset.

\subsection{Image Classification}\label{sec:img_cls}

\begin{table}[!t]
    \centering
    \caption{Results of image classification on ImageNet.}
    \label{tab:imagenet_cls}
\renewcommand\arraystretch{.9}
\setlength{\tabcolsep}{6pt}
\resizebox{\columnwidth}{!}{
    \begin{tabular}{cl|cc|cc|c}
        \toprule
         & \textbf{Method} & \textbf{Base} & \textbf{Pruned} & \textbf{$\Delta$ Acc.} & \textbf{FLOPs} & \textbf{Speedup} \\
        \midrule \midrule
         \multirow{8}{*}{\rotatebox{90}{\textbf{ResNet-50}}}
         & ResNet-50 & 76.34 & - & - & 4.13 & - \\
         & Taylor~\cite{molchanovImportanceEstimationNeural2019} & 76.18 & 74.50 & -1.68 & 2.25 & 1.83 \\
         & GFP~\cite{liuGroupFisherPruning2021} & 76.79 & 76.42 & -0.37 & 2.04 & 2.02 \\
         & AutoSlim~\cite{yuAutoSlimOneShotArchitecture2019} & 76.10 & 75.60 & -0.50 & 2.00 & 2.06 \\
         & SFP~\cite{heSoftFilterPruning2018} & 76.15 & 74.61 & -1.54 & 2.40 & 1.72 \\
         & GReg-2~\cite{wangNeuralPruningGrowing2021} & 76.13 & 75.36 & -0.77 & 2.77 & 1.49 \\
         & DepGraph~\cite{fangDepGraphAnyStructural2023} & 76.15 & 75.83 & -0.32 & 1.99 & 2.07 \\
         &\cellcolor{gray!10}Ours & \cellcolor{gray!10}76.34 & \cellcolor{gray!10}75.91 & \cellcolor{gray!10}-0.43 & \cellcolor{gray!10}1.92 & \cellcolor{gray!10}2.15 \\
         \midrule
         \multirow{5}{*}{\rotatebox{90}{\textbf{MNet V2}}}
         & MobileNet-v2 & 71.02 & - & - & 0.33 & - \\
         & Meta~\cite{liuMetaPruningMetaLearning2019} & 74.70 & 68.20 & -6.50 & 0.14 & 2.35 \\
         & GFP~\cite{liuGroupFisherPruning2021} & 75.74 & 69.16 & -6.58 & 0.15 & 2.2 \\
         & DepGraph~\cite{fangDepGraphAnyStructural2023} & 71.87 & 68.46 & -3.41 & 0.15 & 2.2 \\
         & \cellcolor{gray!10}Ours & \cellcolor{gray!10}71.02 & \cellcolor{gray!10}67.98 & \cellcolor{gray!10}-3.04 & \cellcolor{gray!10}0.13 & \cellcolor{gray!10}2.53 \\
         \midrule
          \multirow{5}{*}{\rotatebox{90}{\textbf{RNext-50}}}
         & ResNext-50 & 77.33 & - & - & 4.27 & - \\
         & SSS~\cite{huangDataDrivenSparseStructure2018} & 77.57 & 74.98 & -2.59 & 2.43 & 1.75 \\
         & GFP~\cite{liuGroupFisherPruning2021} & 77.97 & 77.53 & -0.44 & 2.11 & 2.02 \\
         & DepGraph~\cite{fangDepGraphAnyStructural2023} & 77.62 & 76.48 & -1.14 & 2.09 & 2.04 \\
         & \cellcolor{gray!10}Ours & \cellcolor{gray!10}77.33 & \cellcolor{gray!10}75.92 & \cellcolor{gray!10}-1.41 & \cellcolor{gray!10}0.99 & \cellcolor{gray!10}4.31  \\
    \bottomrule
    \end{tabular}
    }
    \vspace{-0.2cm}
\end{table}

\Cref{tab:imagenet_cls} shows the results of our pruning method compared to previous state-of-the-art approaches. To show the generalizability of our approach, we choose to use several architecturally different models. This includes ResNet~\cite{he2016resnet} for the use of residual connections, MobileNet-v2~\cite{sandlerMobileNetV2InvertedResiduals2019a} for depth-wise convolutions and ResNext~\cite{xieAggregatedResidualTransformations2017} for grouped convolutions. We show that our approach which combines a simple pruning criterion with KD can achieve or surpass current SOTA pruning methods achieving speedups of over to $\times2.15$ compared to their baseline while keeping the accuracy loss at an acceptable level. In the case of ResNext-50 we even manage to reduce the FLOPs by over $\times4.31$ with a relative accuracy loss of only $1.8\%$.

\subsection{Semantic Segmentation}\label{sec:sem_seg}

\begin{table}[!t]
    \centering
    \caption{Results of semantic segmentation on ADE20K.}
    \label{tab:ade20k}
\renewcommand\arraystretch{1.}
\setlength{\tabcolsep}{10pt}
\resizebox{\columnwidth}{!}{
    \begin{tabular}{l|c|c}
        \toprule
         \textbf{Model} & \textbf{mIoU (\%)} & \textbf{Latency (ms)} \\
        \midrule \midrule
         Swin-T~\cite{liu2021swin} & 44.41 & 70.60 \\
         PP-MobileSeg-Base~\cite{tangPPMobileSegExploreFast2023} & 41.57 & 60.13 \\
         ConvNext + UperNet~\cite{liu2022convnext} & 46.11 & 59.60 \\
         DANet (R-50-D8)~\cite{fu2019danet} & 42.90 & 43.83 \\
         FastFCN (R50-D32) + PSP~\cite{wuFastFCNRethinkingDilated2019} & 42.12 & 35.93 \\
         \midrule
         ViT-DeiT-S~\cite{dosovitskiy2021vit} & 42.96 & 18.80 \\
         \rowcolor[gray]{.9}ViT-DeiT-S (Ours) & 40.77 & 9.92 \\
         SeaFormer Large~\cite{wan2023seaformer} & 42.04 & 9.28 \\
         \rowcolor[gray]{.9}SeaFormer Large (Ours) & 41.29 & 6.84 \\
    \bottomrule
    \end{tabular}
    }
    \vspace{-0.1cm}
\end{table}
We evaluate the performance of SeaFormer~\cite{wan2023seaformer} and ViT~\cite{dosovitskiy2021vit} on the ADE20K~\cite{zhou2017ade20k} dataset for semantic segmentation. For all experiments in \Cref{tab:ade20k} we use channel-wise KD~\cite{shuChannelwiseKnowledgeDistillation2021} (CWD) as our KD method. For example, we reduce the measured latency of ViT-DeiT-S by over $47\%$. For the SeaFormer-L we achieve a smaller latency improvement of 
$26\%$ with a even smaller performance loss of $1.7\%$.
The latency is calculated using the exported ONNX model respectively. This is done approximate a real-world usage scenario. Each results is the mean latency over a large number ($n\geq1000$) of evaluations.

\subsection{Ablation Study}\label{sec:ablation}

\begin{figure}
    \centering
    \begin{tikzpicture}[scale=.9]
        \begin{axis}[
            xlabel={Sparsity (\%)},
            ylabel={mIoU (\%)},
            legend pos=south west,
            ymajorgrids=true,
            grid style=dashed,
        ]
        
        \addplot[
            color=blue,
            mark=square,
            ]
            coordinates {
            (0,42.96)(20,41.11)(30,40.35)(40,40.77)(50,34.53)
            };
        \draw (axis cs:40,40.77) node[blue, diamond, fill=blue, draw, scale=0.6] {};
        \addplot[
            color=red,
            mark=o,
            ]
            coordinates {
            (0,42.04)(20,41.29)(30,40.85)(40,40.73)(50,22.91)
            };
        \draw (axis cs:40,40.73) node[red, diamond, fill=red, draw, scale=0.6] {};
        \legend{ViT-DeiT-S, SeaFormer-L};
        \end{axis}
    \end{tikzpicture}
\vskip -1ex
\caption{Sparsity and mIoU of ViT-DeiT-S on ADE20K. }
\label{ablation:spars_perf_trade}
\vskip -2ex
\end{figure}
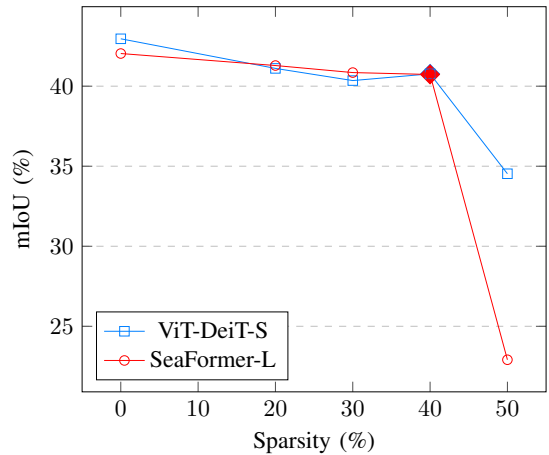
\noindent\textbf{Tradeoff between Sparsity and Performance.} In \Cref{ablation:spars_perf_trade} we present our findings regarding the impact of sparsity on the predictive performance of a model.
The sparsity is calculated as the relative percentage of parameters removed in a model.
For this example we use CWD KD with the ViT-DeiT-S model. In this case, we can observe a expected, relatively stable drop in performance up until the critical point with a sparsity of about 40\%. Afterwards a rather large drop in accuracy occurs. Similar behavior can also be seen with other models although the critical point of sparsity may occur at a lower or higher amount depending on the model. We hypothesize that the critical point may scale with the size of the original model, i.e., larger models tend to have this critical point at a higher sparsity then smaller models.

\begin{table}
    \centering
    \caption{Results for SeaFormer Small using KD with different sized teachers on ADE20K.}
    \label{tab:ablation_seaformer_teachers}
\renewcommand\arraystretch{1.}
\resizebox{\linewidth}{!}{%
\setlength{\tabcolsep}{12pt}
    \begin{tabular}{ll|l|l}
        \toprule
        Teacher & Student & mIoU & $\Delta$~Acc. \\
        \midrule \midrule
        SeaFormer-Large & SeaFormer-Small & 37.36 & -1.05 \\
        SeaFormer-Small & SeaFormer-Small & \textbf{37.73} & \textbf{-0.68}  \\
        \bottomrule
    \end{tabular}
}
\end{table}

\noindent\textbf{Effect of Teacher Selection.} Since many models provide different architecture variants, we conduct tests on the advantage of using larger teachers of the same family instead of the same, unpruned model as the teacher. \Cref{tab:ablation_seaformer_teachers} shows that in our experiments, our method can use the unpruned model instead of a larger teacher. 

\begin{table}[t]
    \centering
    \caption{Effect of using different KD methods on our CPD framework. The mIoU results are evaluated on ADE20K. 
    }
    \label{tab:ablation_seaformer_kd}
\renewcommand\arraystretch{1.}
\setlength{\tabcolsep}{29pt}
\resizebox{\linewidth}{!}{%
    \begin{tabular}{l|ll}
        \toprule
        Method & mIoU & $\Delta$ mIoU \\
        \midrule \midrule
        No KD & 40.46 & -1.58 \\
        KL~\cite{hintonDistillingKnowledgeNeural2015}    & 41.07 & -0.97 \\
        CIRKD~\cite{yangCrossImageRelationalKnowledge2022} & 41.13 & -0.91 \\
        CWD~\cite{shuChannelwiseKnowledgeDistillation2021}   & \textbf{41.29} & \textbf{-0.75}  \\
        \bottomrule
    \end{tabular}
}
\vskip -3ex
\end{table}

\pgfplotstableread[col sep=comma]{miou_scores.csv}\miouclassscores

\makeatletter
\pgfplotsset{
    /pgfplots/flexible xticklabels from table/.code n args={3}{%
        \pgfplotstableread[#3]{#1}\coordinate@table
        \pgfplotstablegetcolumn{#2}\of{\coordinate@table}\to\pgfplots@xticklabels
        \let\pgfplots@xticklabel=\pgfplots@user@ticklabel@list@x
    }
}
\makeatother

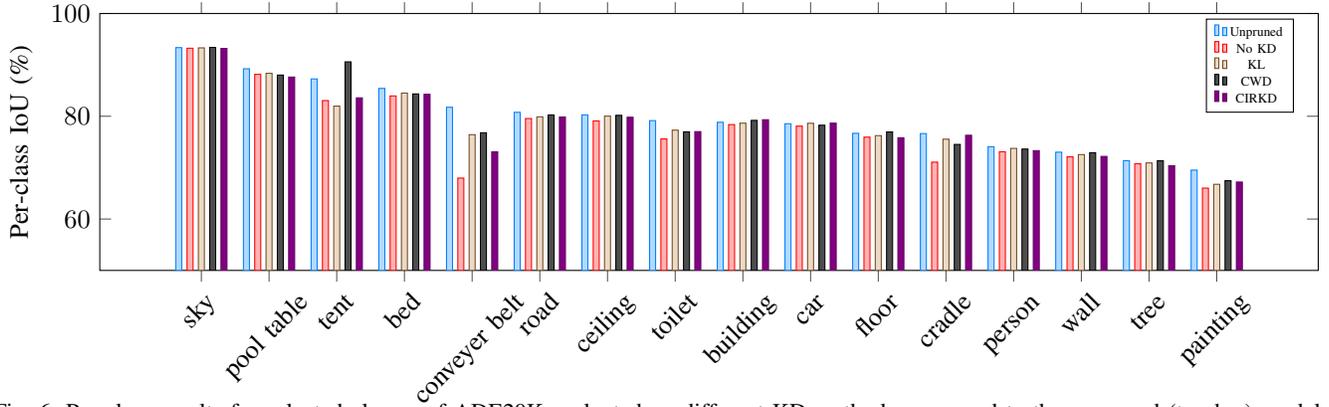
\begin{figure*}
\centering
\begin{tikzpicture}
\begin{axis}[
ybar, ymin=50, ymax=100,
ylabel=Per-class IoU (\%),
width=\linewidth,
height=5cm,
flexible xticklabels from table={miou_scores.csv}{Class}{col sep=comma},
xticklabel style={text height=1ex,rotate=45}, 
bar width=0.08cm,
xtick=data,
legend style={nodes={scale=0.5, transform shape}},
legend image post style={scale=0.5, solid},
]
\addplot table[x expr=\coordindex,y=Unpruned]{\miouclassscores};
\addplot table[x expr=\coordindex,y=NO20]{\miouclassscores};
\addplot table[x expr=\coordindex,y=KL20]{\miouclassscores};
\addplot table[x expr=\coordindex,y=CWD20]{\miouclassscores};
\addplot table[x expr=\coordindex,y=CIRKD20]{\miouclassscores};
\legend{Unpruned, No KD, KL, CWD, CIRKD};

\end{axis}

\end{tikzpicture}
\vskip -3ex
\caption{Per-class results for selected classes of ADE20K evaluated on different KD methods compared to the unpruned (teacher) model.}
\label{fig:classes_miou_scores}
\end{figure*}

\noindent\textbf{Effect of KD Methods.} To test different KD methods, we chose to use SeaFormer-Large as a baseline on the ADE20K dataset. In our experiments in~\Cref{tab:ablation_seaformer_kd} we find that using CWD when pruning retains the most performance of the original model. 
In the context of our pruning experiments we find that CWD performs better than CIRKD. 
We hypothesize that this is because that during pruning we use a already pre-trained model as a starting point and, in contrast to the original works, do not start from a untrained student model. Therefore, the more global semantic information that CIRKD is focused on is already learned by the model at this point and this information is retained in the pruning process while the smaller details are predominantly removed. Due to this, we theorise that during the pruning process the most important part of KD is to make sure that no details are lost.

~\Cref{fig:classes_miou_scores} shows some of the individual class scores of the different KD methods on the ADE20K dataset. Due to the large number of classes we selected the 16 classes with the highest mIoU on the unpruned teacher model. Each included KD method was evaluated at a model sparsity of $20\%$. We can observe that generally pruned models tend to perform similar to their unpruned teacher regardless of KD method. The selection of KD method also does not seem have a large impact on the individual class performance as the class scores are similar in the pruned models. The largest differences compared to the unpruned model can be observed for smaller and more uncommon classes. On those classes we can also measure the largest differences between KD methods. 

\section{Conclusion}
\label{conclusion}
In this work, we propose a novel unified model compression framework called CPD, which includes three major steps: combing, pruning, and distillation. To solve the limitations of previous model-specific pruning methods relying on network architecture, our framework incorporates a dependency-resolving algorithm, ensuring its flexibility to be applied to various architectures. Furthermore, we combine Knowledge Distillation (KD) into the framework to improve the pruned model. 
We show that compared to previous state-of-the-art approaches, using KD in the pruning process can improve the retention of performance when pruning a model. Extensive experiments on two datasets prove the effectiveness of our CPD framework. 

\noindent\textbf{Limitations.} Our proposed pruning framework is currently verified on the image classification and semantic segmentation tasks. There are two different architectures included in the experiments, i.e., the CNN-based and Transformer-based architectures. In the future work, we plan to explore the unified framework in MLP-based architectures and to include more different vision tasks. Furthermore, applying this unified framework to compress large language model or vision-language models is a promising direction.   

\bibliographystyle{IEEEtran}
\bibliography{main}
\end{document}